%% file: main.tex
\documentclass[runningheads]{llncs}
\usepackage{graphicx}
\graphicspath{{figures/}}
\usepackage{amsmath,amssymb} 
\usepackage[usenames]{color}
\usepackage[breaklinks=true,colorlinks,bookmarks=false]{hyperref}
\usepackage[width=122mm,left=12mm,paperwidth=146mm,height=193mm,top=12mm,paperheight=217mm]{geometry}
\usepackage{booktabs}
\usepackage{makecell}
\usepackage{wrapfig}
\usepackage{breakcites}
\usepackage{tabu}
\usepackage{floatrow}

\newcommand{\ra}[1]{\renewcommand{\arraystretch}{#1}}
\newcommand*\samethanks[1][\value{footnote}]{\footnotemark[#1]}
\newcommand{\sus}[1]{\textsuperscript{#1}}

\begin{document}
\title{The 2018 PIRM Challenge on\\Perceptual Image Super-resolution} 

\titlerunning{The 2018 PIRM Challenge on Perceptual Image Super-resolution}

\author{Yochai Blau\inst{1}\thanks{indicates authors who contributed equally.} \and Roey Mechrez\inst{1}\samethanks[1] \and Radu Timofte\inst{2} \and\\Tomer Michaeli\inst{1} \and Lihi Zelnik-Manor\inst{1}}

\authorrunning{Y. Blau*, R. Mechrez*, R. Timofte, T. Michaeli, and L. Zelnik-Manor}

\institute{Technion--Israel Institute of Technology, Haifa, Israel
\and ETH Zurich, Switzerland\\
\email{\{yochai,roey\}@campus.technion.ac.il}}

\maketitle
\setcounter{footnote}{0}

\input{abstract}
\input{intro}
\input{perceptualSR}
\input{challenge}
\input{results}

\input{qualityMeasures}
\input{currentTrends}

\input{conclusions}

\bibliographystyle{splncs04}
\bibliography{PIRM_bib}

\clearpage
\appendix
\input{appendices}

\end{document}

%% file: abstract.tex
\begin{abstract}
This paper reports on the 2018 PIRM challenge on perceptual super-resolution (SR), held in conjunction with the Perceptual Image Restoration and Manipulation (PIRM) workshop at ECCV 2018. In contrast to previous SR challenges, our evaluation methodology jointly quantifies \emph{accuracy} and \emph{perceptual quality}, therefore enabling perceptual-driven methods to compete alongside algorithms that target PSNR maximization. Twenty-one participating teams introduced algorithms which well-improved upon the existing state-of-the-art methods in perceptual SR, as confirmed by a human opinion study. We also analyze popular image quality measures and draw conclusions regarding which of them correlates best with human opinion scores. We conclude with an analysis of the current trends in perceptual SR, as reflected from the leading submissions.
\end{abstract}

%% file: intro.tex
\section{Introduction}\label{sec:intro}
\begin{figure}[t]
	\begin{center}
		\includegraphics[width=1\linewidth]{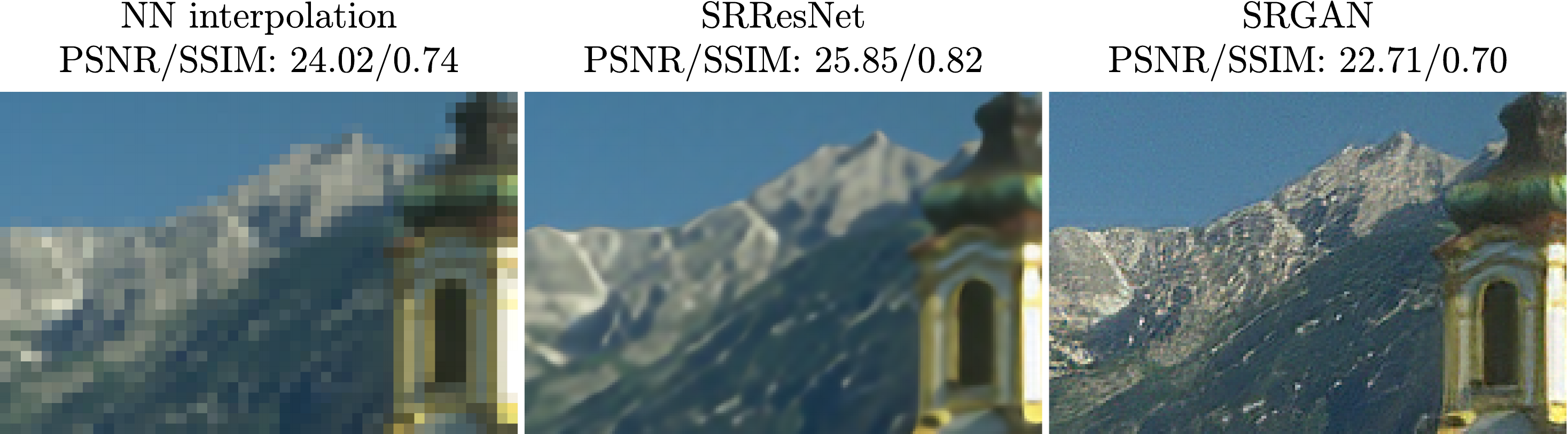}
		\caption{\textbf{Inconsistency between PSNR\slash SSIM values and perceptual quality.} From left to right: nearest-neighbor (NN) interpolation, SRResNet \cite{ledig2017photo} which aims for high PSNR, and SRGAN \cite{ledig2017photo} which aims for high perceptual quality. The perceptual quality of SRGAN is far better than SRResNet. However, its PSNR\slash SSIM values are substantially lower than those of SRResNet, and even lower than those of NN interpolation. The image is from the BSD dataset \cite{martin2001database}.}\label{fig:SRexample}
	\end{center}
\end{figure}

The past few years have seen a major performance leap in single-image super-resolution (SR), both in terms of reconstruction accuracy (as measured e.g.,~by PSNR, SSIM) \cite{lim2017enhanced,haris2018deep,tong2017image,timofte2017ntire,timofte2018ntire} and in terms of visual quality (as rated by human observers) \cite{ledig2017photo,mechrez2018learning,sajjadi2017enhancenet,wang2018recovering,wang2018fully}. However, the more SR methods advanced, the more it has become evident that reconstruction accuracy and perceptual quality are typically in disagreement with each other. That is, models which excel at minimizing the reconstruction error tend to produce visually unpleasing results, while models that produce results with superior visual quality are rated poorly by distortion measures like PSNR, SSIM, IFC, etc. \cite{ledig2017photo,johnson2016perceptual,mechrez2018learning,sajjadi2017enhancenet,dahl2017pixel} (see Fig.~\ref{fig:SRexample}). Recently, it has been shown that this disagreement cannot be completely resolved by seeking for better distortion measures \cite{blau2017perception}. Namely, there is a fundamental tradeoff between the ability to achieve low distortion and low deviation from natural image statistics, no matter what full-reference dissimilarity criterion is used to measure distortion.

\begin{figure}[t]
	\begin{center}
		\includegraphics[width=0.65\linewidth]{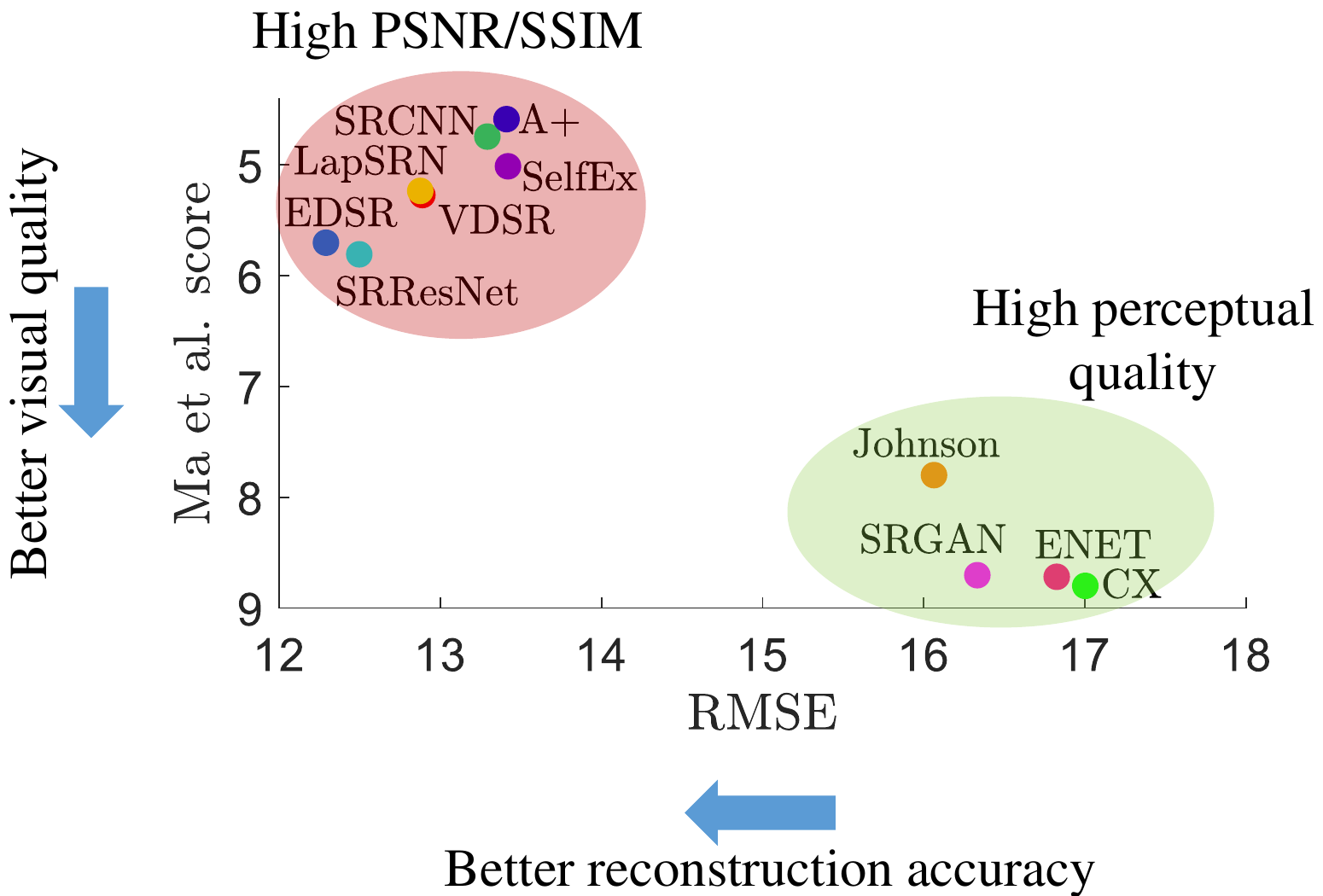}
		\caption{\textbf{Two directions in image super-resolution.} Super-resolution algorithms, plotted according to the mean reconstruction accuracy (measured by RMSE values) and mean perceptual quality (measured by the recent metric \cite{ma2017learning}). Current methods group into two clusters: (i--\emph{upper-left}) high PSNR\slash SSIM and (ii--\emph{lower-right}) high perceptual quality. Scores are computed on the BSD test set \cite{martin2001database}. The plotted methods are \cite{timofte2014a+,huang2015single,dong2014learning,kim2016accurate,lai2017deep,lim2017enhanced,ledig2017photo,mechrez2018learning,sajjadi2017enhancenet,johnson2016perceptual}.}\label{fig:PD_plot}
	\end{center}
\end{figure}

These observations caused the formation of two distinct research trends (see Fig.~\ref{fig:PD_plot}). The first is aimed at improving the reconstruction accuracy according to popular full-reference distortion metrics, and the second targets high perceptual quality. While reconstruction accuracy can be precisely quantified, perceptual quality is often estimated through user studies, in which, due to practical limitations, each user is typically exposed to only a small number of methods and/or a small number of images per method. Therefore, reports on perceptual quality are often inaccurate and hard to reproduce. As a result, novel methods cannot be easily compared to their predecessors in terms of perceptual quality, and existing benchmarks and challenges (e.g.,~NTIRE \cite{timofte2018ntire}) focus mostly on quantifying reconstruction accuracy, using e.g.,~PSNR\slash SSIM. As perceptually-aware super-resolution is gaining increasing attention in recent years, there is a need for a benchmark for evaluating perceptual-quality driven algorithms.

The 2018 PIRM challenge on perceptual super-resolution took part in conjunction with the 2018 Perceptual Image Restoration and Manipulation (PIRM) workshop. This challenge compared and ranked \emph{perceptual} super-resolution algorithms. In contrast to previous challenges, the evaluation was performed in a perceptual-quality aware manner, as suggested in \cite{blau2017perception}. Specifically, we define perceptual quality as the visual quality of the reconstructed image \emph{regardless} of its similarity to any ground-truth image. Namely, it is the extent to which the reconstruction looks like a valid natural image. Therefore, we measured the perceptual quality of the reconstructed images using \emph{no}-reference image quality measures, which do not rely on the ground-truth image.

Although the main motivation of the challenge is to promote algorithms that produce images with good perceptual quality, similarity to the ground truth images is obviously also of importance. For example, perfect perceptual quality can be achieved by randomly drawing natural images that have nothing to do with the input images. Such a scheme would score quite poorly in terms of reconstruction accuracy. We therefore evaluate algorithms on a {2-dimensional} plane, where one axis is the full-reference root mean squared error (RMSE) distortion, and the second axis is a perceptual index which combines the \emph{no}-reference image quality measures of \cite{mittal2013making} and~\cite{ma2017learning}. This approach jointly quantifies \emph{accuracy} and \emph{perceptual} quality, thus enabling perceptual-driven methods to compete alongside algorithms that target PSNR maximization. PIRM is therefore the first established benchmark for perceptual-quality driven image restoration, which will hopefully be extended to other perceptual computer-vision tasks in the future. 
\\

\noindent The outcomes arising from this challenge are manifold:

\noindent  $\bullet$ Participants introduced algorithms which well-improve upon the state of the art in perceptual SR. The submitted methods incorporated novelties in optimization objectives (losses), conv-net architectures, generative adversarial net (GAN) variants, training schemes and more. These enabled to impressively surpass the performance of baselines, such as EnhanceNet \cite{sajjadi2017enhancenet} and CX \cite{mechrez2018learning}. The results are presented in Section \ref{sec:results}, and the main novelties are discussed in Section~\ref{sec:trends}.

\noindent  $\bullet$ We validate our chosen perceptual index through a human-opinion study, and find that it is highly correlated with the ratings of human observers. This provides empirical evidence that no-reference image quality measures can faithfully assess perceptual quality. The results of the human-opinion study are presented in Section \ref{sec:humanStudy}.

\noindent  $\bullet$ We also test the agreement of many other commonly used image quality measures with the human-opinion scores, and find that \emph{most} of them are either uncorrelated or \emph{anti}-correlated. This shows that most existing schemes for evaluating image restoration algorithms cannot be used to quantify perceptual quality. The results of this analysis are presented in Section~\ref{sec:qualityMeasures}.

\noindent  $\bullet$ The challenge results provide insights on the trade-off between perception and distortion (suggested and analyzed in \cite{blau2017perception}). In particular, at the low-distortion regime, participants showed considerable improvements in perceptual quality over methods that excel in RMSE (e.g.~EDSR \cite{lim2017enhanced}), while sacrificing only a small increase in RMSE. This indicates that the tradeoff is severe in this regime. Furthermore, at the good perceptual quality regime, participants were able to improve both in perceptual quality and in distortion, over state-of-the-art perceptual SR methods (e.g.E-Net~\cite{sajjadi2017enhancenet}). This indicates that previous methods were quite far from the theoretical perception-distortion bound discussed in~\cite{blau2017perception}.

%% file: perceptualSR.tex
\section{Perceptual Super Resolution}
The field of image super-resolution (SR) has been dominated by convolutional-network  based methods in recent years. At first, the adopted optimization objective was an $\ell_1\slash \ell_2$ loss, which aimed to improve the reconstruction accuracy (in terms of e.g.~PSNR, SSIM). While the first attempt to apply a conv-net to image SR \cite{dong2014learning} did not significantly surpass the performance of prior methods, it set the ground for major improvements in PSNR\slash SSIM values over the course of the several following years \cite{kim2016accurate,lai2017deep,ledig2017photo,lim2017enhanced,tong2017image,haris2018deep,zhang2018residual,shocher2017zero,han2018image,zhang2018image}. During these years, the rising PSNR\slash SSIM values were not always accompanied by a rise in the perceptual quality. In fact, this resulted in increasingly blurry and unnatural outputs in many cases. These observations led to a significant shift of the optimization objective, from PSNR maximization to perceptual quality maximization. We refer to this new line of works as \emph{perceptual} SR.

The first work to adopt such an objective for SR was that by Johnson et al.~\cite{johnson2016perceptual}, which added an $\ell_2$ loss \emph{on the deep features} extracted from the outputs (commonly referred to as the perceptual loss). The next major breakthrough in perceptual SR was presented by Ledig et al.~\cite{ledig2017photo}, who adopted the perceptual loss and combined it with an adversarial loss (originally suggested for generative modeling by \cite{goodfellow2014generative}). This was further developed in \cite{sajjadi2017enhancenet}, where a texture matching loss was added to the perceptual and adversarial losses. Recently, \cite{mechrez2018learning} showed that natural image statistics can be maintained by replacing the perceptual loss with the contextual loss \cite{mechrez2018contextual}. These ideas were further extended in e.g.,~\cite{wang2018recovering,wang2018fully,gondal2018unreasonable,sun2017super}.

These perceptual SR methods have established a fresh research direction which is producing algorithms with superior perceptual quality. However, in all works, this has come at the cost of a substantial decrease in PSNR and SSIM values, indicating that these common distortion measures do not faithfully quantify the perceptual quality of SR methods \cite{blau2017perception}. As such, perceptual SR algorithms cannot participate in any challenge or benchmark based on these standard measures (e.g.,~NTIRE \cite{timofte2018ntire}), and cannot be compared or ranked using these common metrics.

%% file: challenge.tex
\section{The PIRM Challenge on Perceptual SR}
The PIRM challenge is the first to compare and rank \emph{perceptual} image super-resolution. The essential difference compared to previous challenges is the novel evaluation scheme which is not based solely on common distortion measures such as PSNR\slash SSIM.

\paragraph{\textbf{\emph{Task}}}
The challenge task is $4\times$ super-resolution of a single image which was down-sampled with a bicubic kernel.

\paragraph{\textbf{\emph{Datasets}}}
Validation and testing of the submitted methods were performed on two sets of 100 images each\footnote{The validation set was used throughout the challenge for model development, and the test set was released a week before the challenge ended for assessing the final results.}. These images cover diverse contents, including people, objects, environments, flora, natural scenery, etc. Participants did not have access to the high-res ground truth images during the challenge, and these images were not available on any online source prior to the challenge. These image sets (high and low resolution) are now available online\footnote{\url{https://pirm.github.io}}. Datasets for model training were chosen by the participants.

\begin{figure}[t]
	\begin{center}
		\includegraphics[width=0.6\linewidth]{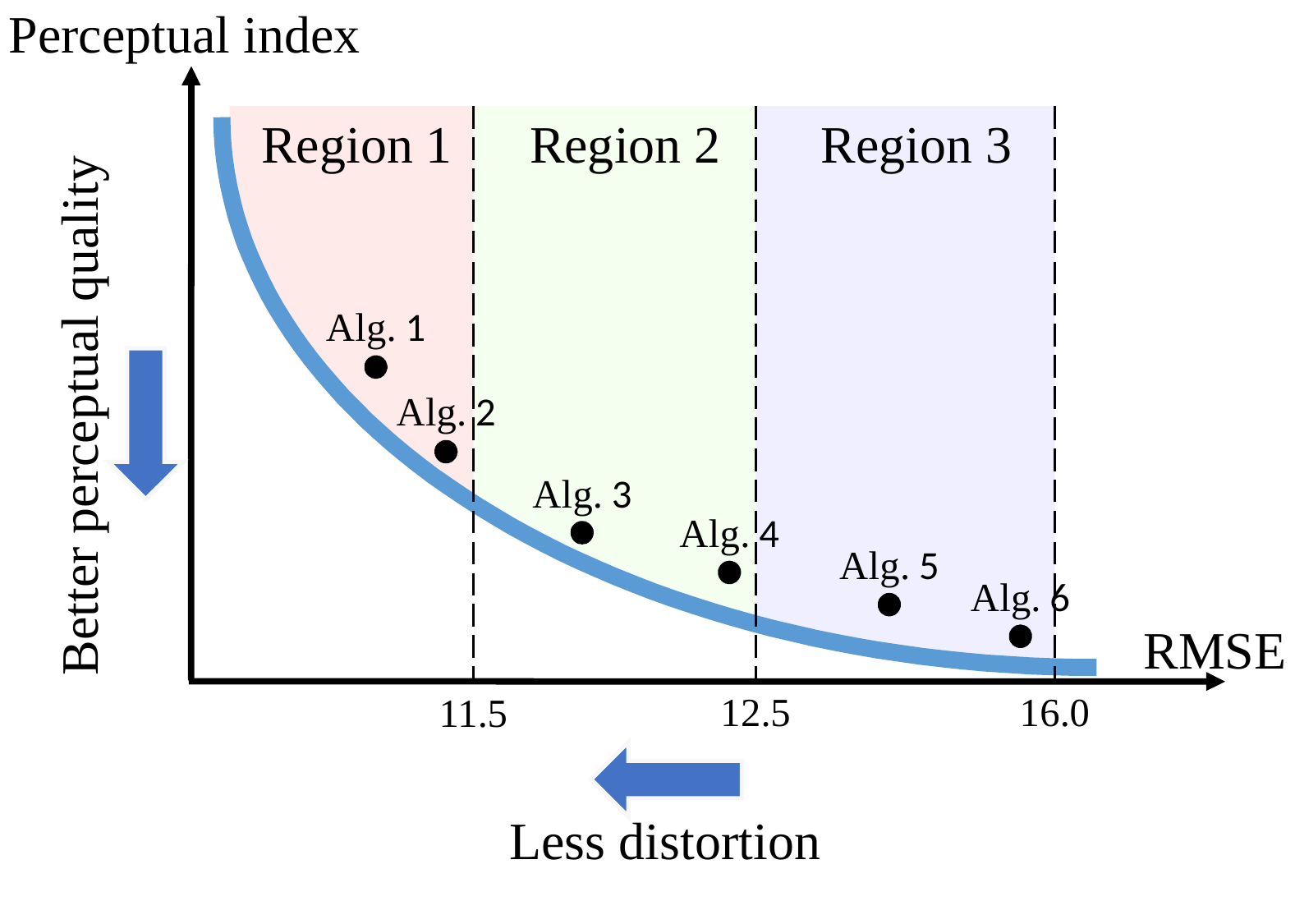}
		\caption{\textbf{Evaluating algorithms on the perception-distortion plane.} 
        The performance of each algorithm is quantified by two measures: (i) the RMSE distortion ($x$-axis), and (ii) the perceptual index, which is based on no-reference image quality measures ($y$-axis, see Eq. \eqref{eq:PI}). It has been shown in \cite{blau2017perception} that the best attainable perceptual quality improves as the allowable distortion level increases (blue curve). In the PIRM challenge, the perception-distortion plane was divided into three regions by placing thresholds on the RMSE. In each region, the challenge goal was to obtain the best perceptual quality.}
\label{fig:PD_plane}
	\end{center}
\end{figure}

\paragraph{\textbf{\emph{Evaluation}}}
The evaluation scheme is based on \cite{blau2017perception}, which proposed to evaluate image restoration algorithms on the perception-distortion plane (see Fig.~\ref{fig:PD_plane}). The rationale of this method is shortly explained in the Introduction.

In the PIRM challenge, the perception-distortion plane was divided into three regions by setting thresholds on the RMSE values (regions $1/2/3$ were defined by $\text{RMSE} \le 11.5/12.5/16$ respectively, see Fig.~\ref{fig:PD_plane}). In each region, the goal was to obtain the best mean perceptual quality. That is, participants attempted to move as downwards as possible in the perception-distortion plane. The perception index (PI) we chose for the vertical axis combines the no-reference image quality measures of Ma et al.~\cite{ma2017learning} and NIQE \cite{mittal2013making} as
\begin{equation}\label{eq:PI}
\text{PI} = \tfrac{1}{2} \left( (10-\text{Ma}) + \text{NIQE} \right).
\end{equation}
Notice that in this setting, a lower perceptual index indicates better perceptual quality. The RMSE was computed as the square-root of the mean-squared-error (MSE) of all pixels in all images\footnote{Note that this is not the mean of the RMSEs of the images, but rather the square-root of the images' mean MSE.}, that is 
\begin{equation}\label{eq:RMSE}
\text{RMSE} = \Big(\tfrac{1}{M} \sum_{i=1}^{M} \tfrac{1}{N_i} \| x_i^{\text{HR}} - x_i^{\text{EST}}\|^2 \Big)^{1/2},
\end{equation}
where $x_i^{\text{HR}}$ and $x_i^{\text{EST}}$ are the $i$th ground truth and estimated images respectively, $N_i$ is the number of pixels in $x_i^{\text{HR}}$, and $M$ is the number of images in the test set. Both the RMSE and the PI were computed on the y-channel after removing a $4$-pixel border. We encouraged participants to submit methods for all three regions, and indeed many did (see Table \ref{tab:results}).

%% file: results.tex
\section{Challenge Results}\label{sec:results}
\begin{figure}
	\begin{center}
		\includegraphics[width=0.8\linewidth]{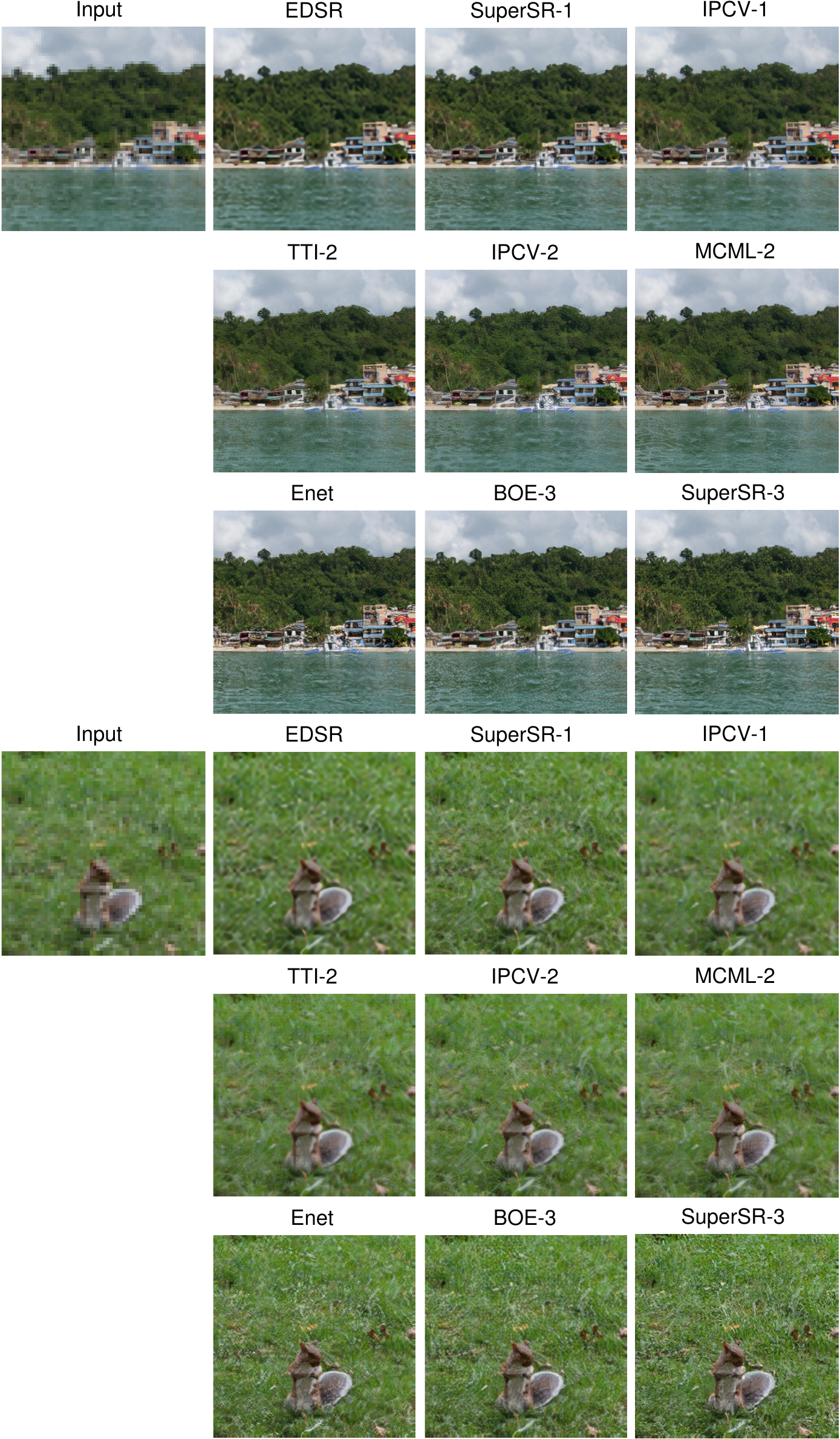}
		\caption{\textbf{Visual results.} SR results of several top methods in each region, along with the EDSR \cite{lim2017enhanced} and EnhanceNet \cite{sajjadi2017enhancenet} baselines. The attainable perceptual quality becomes higher as the allowed RMSE increases.}\label{fig:topImages}
	\end{center}
\end{figure}

Twenty-one teams participated in the test phase of the challenge. Table \ref{tab:results}
reports the top scoring teams in each region, where the team members and affiliations can be found in Appendix \ref{ap:teamMembers}. Figure \ref{fig:testGraph}(a) plots all test phase submissions on the perception-distortion plane (teams were allowed up to 10 final submissions). Figure \ref{fig:testGraph}(b) shows the correlation between our perceptual index (PI) and human-opinion-scores on the top 10 submissions (see details in Sec.~\ref{sec:qualityMeasures}). The high correlation justifies our choice of definition of the PI. In Fig.~\ref{fig:topImages} we compare the visual outputs of several top methods in each region (the number in the method's name indicates the region of the submission), where additional visual comparisons can be found in Appendix \ref{app:moreResults}.  A table with the scores of all participating teams in each region can be found in Appendix \ref{app:fullResults}.

\begin{table}[t]
	\scriptsize
	\centering
	\ra{1.3}
	\begin{tabu}{@{}llcccllcccllcc@{}}\toprule
		\multicolumn{4}{c}{\textbf{Region 1}} & \phantom{abc}& \multicolumn{4}{c}{\textbf{Region 2}} & \phantom{abc} & \multicolumn{4}{c}{\textbf{Region 3}}\\
		\cmidrule{1	- 4} \cmidrule{6 - 9} \cmidrule{11 - 14} \# & Team & \makecell{PI} & RMSE && \# & Team & \makecell{PI} & RMSE && \# &Team & \makecell{PI} & RMSE\\
		\cmidrule{1	- 4} \cmidrule{6 - 9} \cmidrule{11 - 14} 
		\rowfont{\color{blue}}	$1$  & IPCV \cite{vasu2018analyzing} & 2.709 & 11.48 && $1$ 	& TTI			& 2.199	& 12.40	&& $1$	& SuperSR \cite{wang2018esrgan}		& 1.978 & 15.30\\
		\rowfont{\color{blue}}	$2$  & MCML \cite{cheon2018generative} & 2.750	& 11.44	&& $2*$ & IPCV \cite{vasu2018analyzing}		& 2.275 & 12.47	&& $2$  & BOE \cite{Navarrete2018multi}			& 2.019	& 14.24\\
		\rowfont{\color{blue}}	$3*$ & SuperSR \cite{wang2018esrgan} & 2.933	& 11.50	&& $2*$ & MCML \cite{choi2018deep} & 2.279 & 12.41 && $3$  & IPCV \cite{vasu2018analyzing}		& 2.013	& 15.26\\
		\color{blue}{$3*$} & \color{blue}{TTI} & \color{blue}{2.938} & \color{blue}{11.46} && $4$ & SuperSR	\cite{wang2018esrgan} & 2.424	& 12.50	&& $4$  & AIM \cite{vu2018perception} & 2.013 & 15.60\\
								$5$  & AIM \cite{vu2018perception}			& 3.321	& 11.37	&& $5$ 	& BOE \cite{Navarrete2018multi}			& 2.484 & 12.50	&& $5$  & TTI			& 2.040	& 13.17\\
								$6$  & DSP-whu		& 3.728	& 11.45	&& $6$ 	& AIM \cite{vu2018perception}			& 2.600 & 12.42	&& $6$  & Haiyun \cite{luo2018bi} & 2.077	& 15.95\\
								$7*$ & BOE \cite{Navarrete2018multi} & 3.817	& 11.50	&& $7$ 	& REC-SR \cite{kuldeep2018scale} & 2.635 & 12.37	&& $7$  & gayNet		& 2.104	& 15.88\\
								$7*$ & REC-SR \cite{kuldeep2018scale}		& 3.831	& 11.46	&& $8$ 	& DSP-whu		& 2.660 & 12.24 && $8$  & DSP-whu		& 2.114	& 15.93\\
								$9$  & Haiyun \cite{luo2018bi}	& 4.440	& 11.19	&& $9$ 	& XYN			& 2.946 & 12.23	&& $9$  & MCML	& 2.136	& 13.44\\
		\bottomrule
	\end{tabu}
	\caption{\textbf{Challenge results.} The top 9 submissions in each region. For submissions with a marginal PI difference (up to 0.01), the one with the lower RMSE is ranked higher. Submission with marginal differences in both the PI and RMSE are ranked together (marked by $*$). We perform a human-opinion-study on the {\color{blue}top submissions} colored in blue (see Section \ref{sec:humanStudy}). See the cited papers describing the submissions. Team members and affiliations can be found in Appendix \ref{ap:teamMembers}. A full table of the test phase results appears in Appendix \ref{app:fullResults}.}\label{tab:results}
\end{table}

\begin{figure}[t]
	\begin{center}
		\includegraphics[width=1\linewidth]{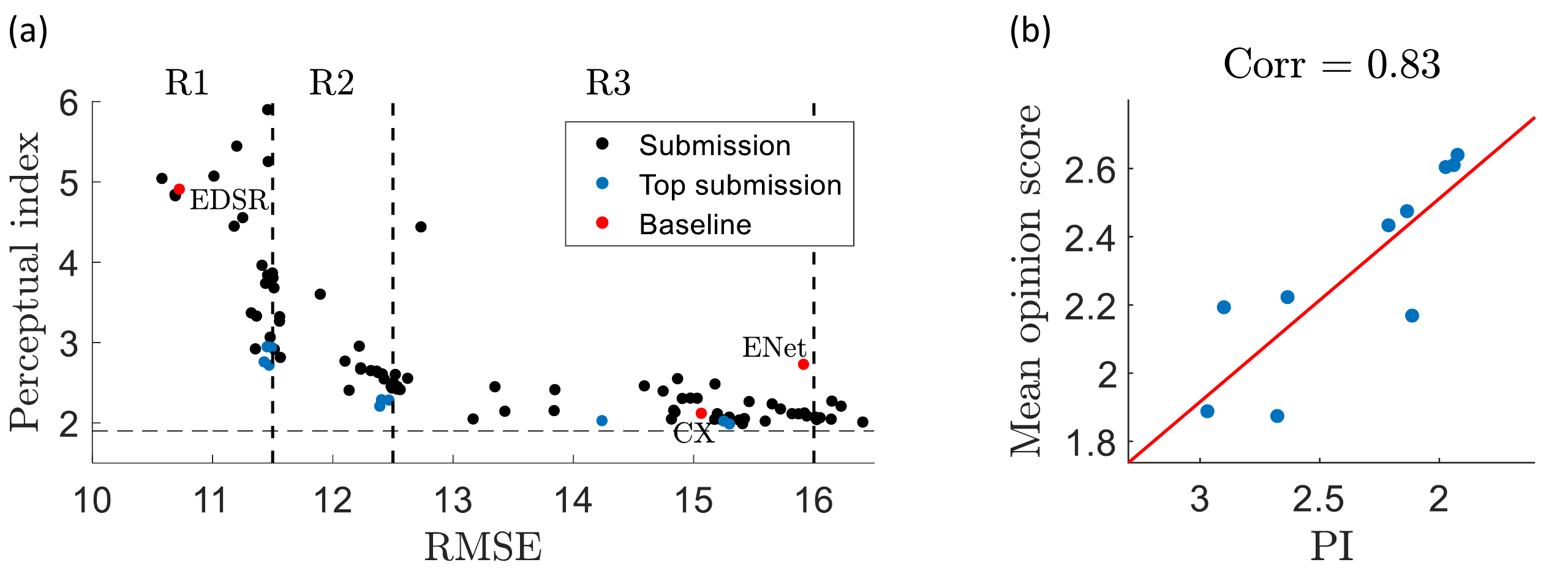}
		\caption{\textbf{Submissions on the perception-distortion plane.} 
        (a) Each submission is a point on the perception-distortion plane, whose axes are RMSE \eqref{eq:RMSE} and the PI \eqref{eq:PI}. The perceptual quality of the challenge submissions exceeds that of the EDSR \cite{lim2017enhanced}, EnhanceNet \cite{sajjadi2017enhancenet} and CX \cite{mechrez2018learning} baselines (plotted in red). Notice the tradeoff between perceptual quality and distortion, i.e. as the perceptual quality of the submissions improved (lower PI), their RMSE increased. (b) The mean-opinion score of 35 human raters vs. the mean perceptual index (PI) on the 10 top submissions. The PI is highly-correlated with human opinion scores (Spearman’s correlation of 0.83), as visualized by the least squares fit. This validates our choice of definition of the PI. A thorough analysis of other images quality measures appears in Section~\ref{sec:qualityMeasures}.}
        \label{fig:testGraph}
	\end{center}
\end{figure}

The submitted algorithms exceed the performance of previous SR methods in all regions, pushing forward the state-of-the-art in perceptual SR. In Region~$3$, challenge submissions outperform  the EnhanceNet \cite{sajjadi2017enhancenet} baseline, as well as the recently proposed CX \cite{mechrez2018learning} algorithm. Notice that several submissions improve upon the baselines in \emph{both} perceptual quality and reconstruction accuracy, which are both important.  In Region $2$, the top submissions present fairly good perceptual quality with a \emph{far} lower distortion than the methods in Region $3$. Such methods could prove advantageous in applications where reconstruction accuracy is valuable. Inspection of the Region $1$ results reveals that participants obtained a significant improvement in the PI ($45\%$) w.r.t.\@ the EDSR baseline \cite{lim2017enhanced} with only a small increase in the RMSE ($7\%, 0.77$ gray-levels per-pixel).

The results provide insights on the tradeoff between perceptual quality and distortion, which is clearly noticed when progressing from Region $1$ to Region $3$. First, the tradeoff appears to be stronger in the low distortion regime (Region $1$), implying that PSNR maximization can have damaging effects in terms of perceptual quality. In the high perceptual quality regime (Region $3$), notice that beyond some point, increasing the RMSE allows only slight improvement in the perceptual quality. This indicates that it is possible to achieve perceptual quality similar to that of the current state-of-the-art methods with considerably lower RMSE values.

\subsection{Human opinion study}\label{sec:humanStudy}
\begin{figure}[t]
	\begin{center}
		\includegraphics[width=0.65\linewidth]{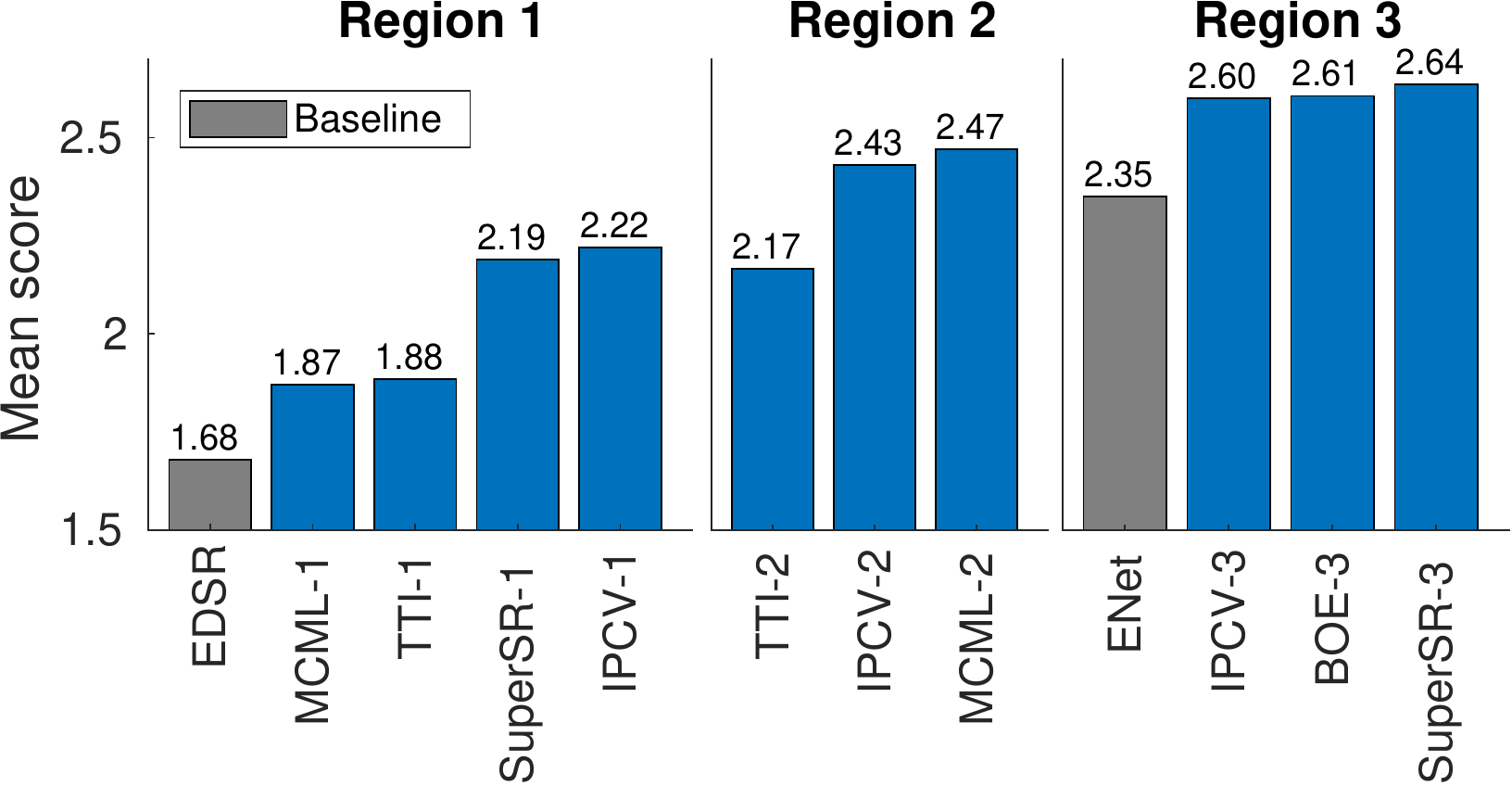}
		\caption{\textbf{Human opinion scores.} 
        Thirty-five human raters rated 12 methods (10 top submissions, 2 baselines). The voting scale was between $1-4$ corresponding to: $1$-Definitely fake, $2$-Probably fake, $3$-Probably real, and 4-Definitely real. These scores validate that the challenge submissions surpassed the performance of state-of-the-art baselines by significant margins. Furthermore, this study shows again that improved perceptual quality can be attained only when allowing higher RMSE values (progressing from region $1$ to $3$).}
        \label{fig:userStudy}
	\end{center}
\end{figure}

We validate the challenge results with a human-opinion study. Thirty-five raters were each shown the outputs of 12 algorithms (10 top challenge submissions, 2~baselines) on 20 images (240 images per rater). For each image, they were asked to rate how realistic the image looked on a scale of $1-4$ which corresponds to: $1$-Definitely fake, $2$-Probably fake, $3$-Probably real, and $4$-Definitely real. We made it clear that ``real'' corresponds to a natural image and ``fake'' corresponds to the output of an algorithm. This scale tests how natural the outputs look. Note that users were not exposed to the original ``ground truth" images, therefore this study does not test distortion in any way, but rather only perceptual quality. The mean human-opinion-scores are shown in Fig.~\ref{fig:userStudy}.

The human-opinion study validates that the challenge submissions surpassed the performance of state-of-the-art baselines by significant margins. Region $3$ submissions, and even Region $2$ submissions, are considered notably better than EnhanceNet by human raters. Region $1$ submissions were rated far better in visual quality compared to EDSR (with only a slight increase in RMSE). The tradeoff between perceptual quality and distortion is once more revealed, as the best attainable perceptual quality increases with the increase in RMSE. Note that while the PI is well correlated with the human-opinion-scores on a coarse scale (in between regions), it is not always well-correlated with these scores on a finer scale (rankings within the regions), which can be seen when comparing the rankings in Table \ref{tab:results} and Fig.~\ref{fig:userStudy}. This highlights the urgent need for better perceptual quality metrics, a point which is further analyzed in Section \ref{sec:qualityMeasures}.

\begin{figure}
\floatbox[{\capbeside\thisfloatsetup{capbesideposition={left,top},capbesidewidth=5.9cm}}]{figure}[\linewidth]
{\caption{\textbf{Human-opinion histogram.} 
Normalized histogram of votes per method. Mean scores are shown as red dots. Notice that all methods fail to achieve a large percentage of ``definitely real'' votes, indicating that there is still much to be done in perceptual super-resolution.}
\label{fig:userStudyHist}}
{		\includegraphics[width=1\linewidth]{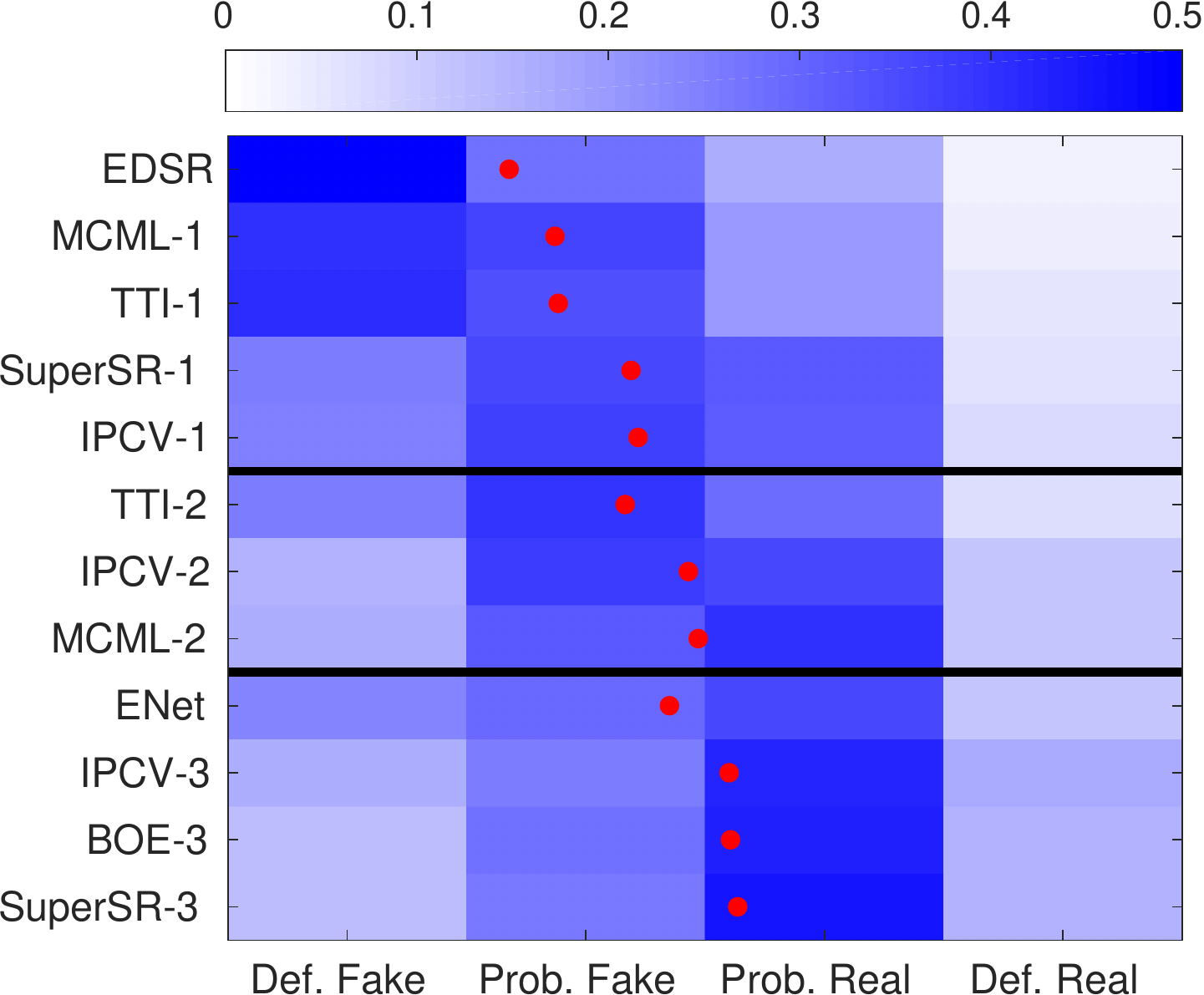}}
\end{figure}

Figure \ref{fig:userStudyHist} shows the normalized histogram of votes per method. Notice that all methods fail to achieve a large percentage of ``definitely real'' votes, indicating that there is still much to be done in perceptual super-resolution. 
In all submitted results, there tend to appear unnatural features in the reconstructions (at $4\times$ magnification), which degrade the perceptual quality. Notice that the outputs of EDSR, a state-of-the-art algorithm in terms of distortion, are mostly voted as ``definitely fake''. This is due to the aggressive averaging causing blurriness as a consequence of optimizing for distortion.

\subsection{Not all images are created equal}\label{sec:variability}
\begin{figure}[t]
	\begin{center}
		\includegraphics[width=1\linewidth,trim={0 1cm 0 0},clip]{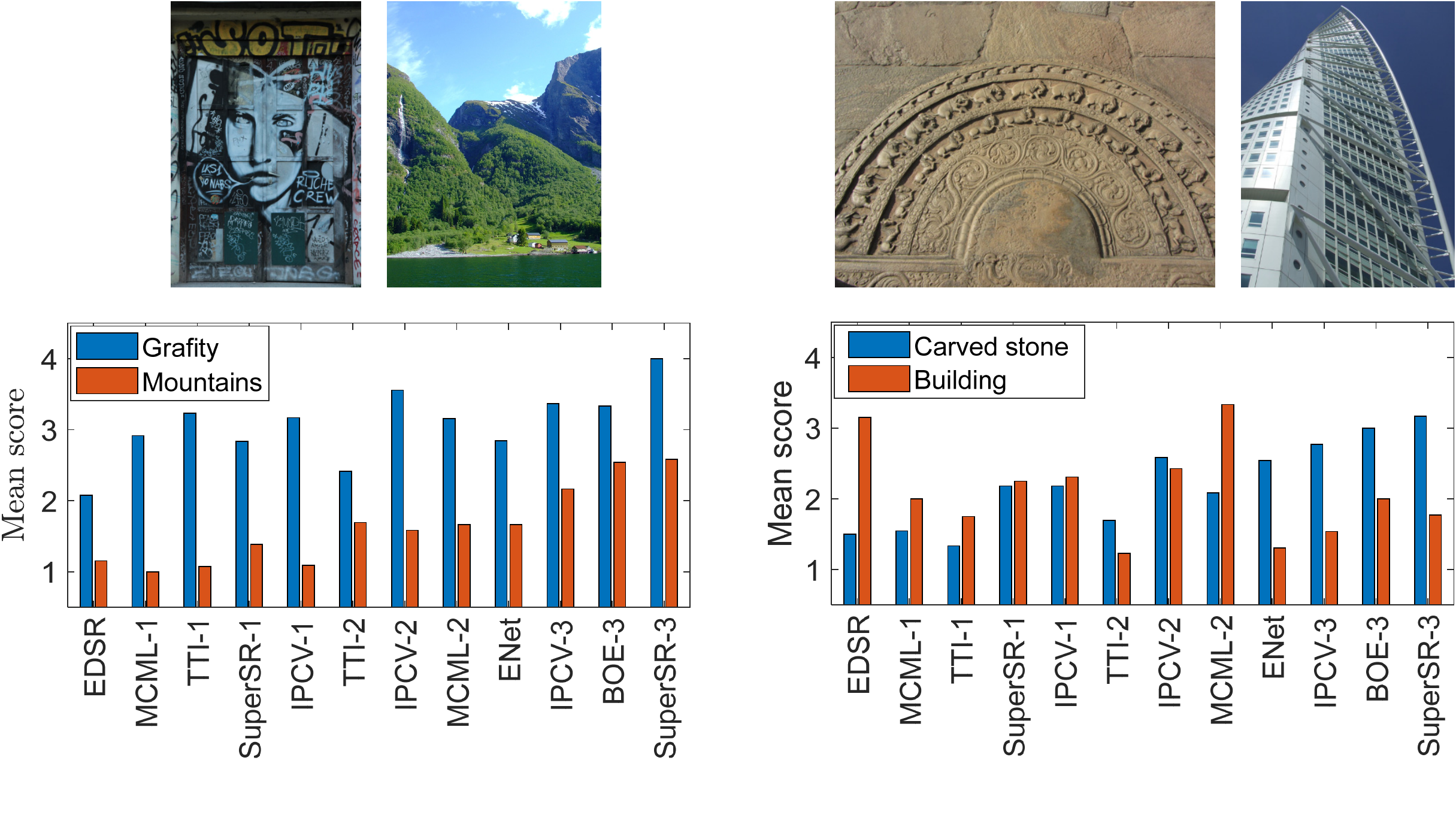}
		\caption{\textbf{Variability between images.} \emph{Left:} Some images are easier to super-resolve than others, where \emph{all} SR methods tend towards high perceptual quality. \emph{Right:} Images dominated by structure are better reconstructed by methods which target accuracy (e.g.~EDSR), while texture-rich images with fine details are reconstructed with high perceptual quality by methods in region $3$.}\label{fig:variability}
	\end{center}
\end{figure}

The results presented in the previous sections show the general trends when averaging over a set of images. Interestingly, when examining single images, there can be quite a variability in SR results. First, there are images which are much easier to super-resolve than others. In such a scenario, the outputs of \emph{all} SR methods tend towards high perceptual quality. Such an example can be seen on the left side of Fig.~\ref{fig:variability}, where the outputs of all methods on the ``grafity'' image are rated fairly higher compared to the ``mountain'' image. In both it seems advantageous to move towards region $3$, but the SR of texture-less images (such as ``grafity'') will generally produce visually pleasing results. Another variation from the average trend are images which include more structure than texture. On such images, it seems that methods from region $1$ which prefer accuracy succeed in maintaining large-scale structures, as opposed to generative-based methods from region~$3$ which tend to distort structures and often produce visually unpleasing results. For example, on the ``building'' image on the right side of Fig.~\ref{fig:variability}, the outputs of EDSR are visually pleasing while the outputs of region~$3$ methods are rated unsatisfactory. However, for images with fine unstructured details such as the ``carved stone'' image, it is beneficial to move towards region $3$. This calls for novel methods, which can either adaptively favor structure preservation vs. texture reconstruction, or employ generative models capable of outputing large-scale structured regions.

%% file: qualityMeasures.tex
\section{Analyzing Quality Measures}\label{sec:qualityMeasures}
\begin{figure}[t]
	\begin{center}
		\includegraphics[width=1\linewidth]{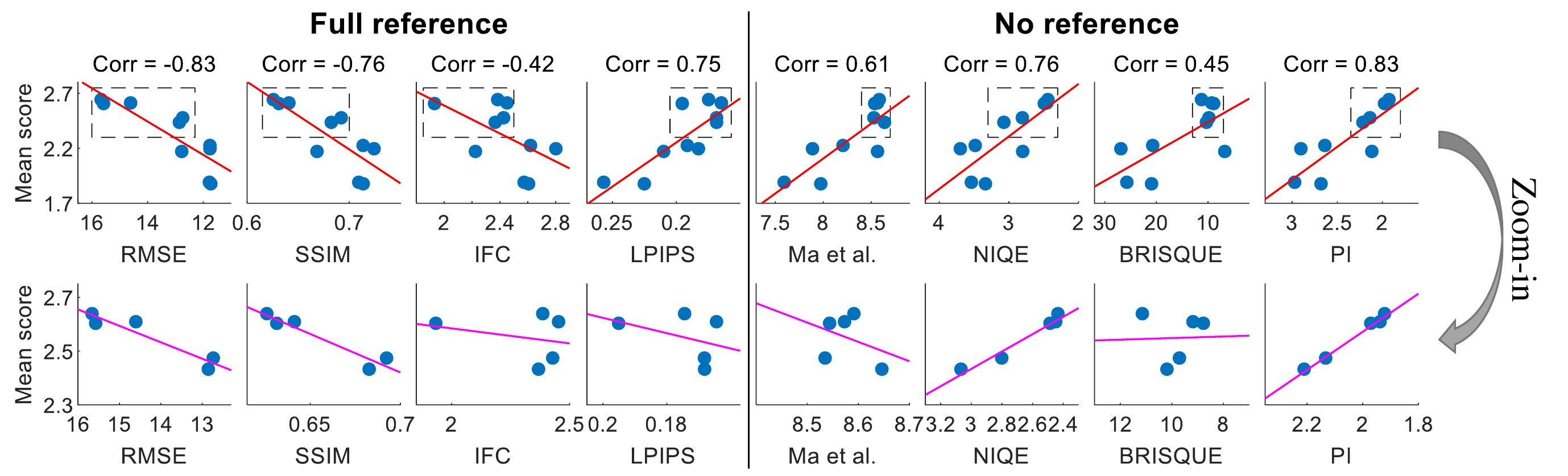}
		\caption{\textbf{Analysis of image quality measures.} 
        \emph{First row:} Scatter plots of mean-opinion-score ($y$-axis) vs. common image quality measures ($x$-axis) for the 10 top challenge submissions, along with Spearman’s correlation coefficients (Corr) and a least-squares linear fit (in red). Note that RMSE, SSIM and IFC are \emph{anti}-correlated with human-opinion-scores, and that our PI is the most correlated. Second row: zoom-in on the high perceptual quality regime (mean scores above $2.3$), and the corresponding least-squares linear fits in magenta. In this regime, even the LPIPS, Ma, and BRISQUE measures, which score well on the first row, do not correlate with the human raters' scores and only NIQE and our PI have high correlations.} 
        \label{fig:corrMethods}
	\end{center}
\end{figure}

The lack of a faithful criterion for assessing the perceptual quality of images is restricting progress in perceptually-aware image reconstruction and manipulation tasks. The current main tool for comparing methods are human-opinion studies, which are hardly reproducible, making it practically impossible to systematically compare methods and assess progress. Here, we analyze the relation between existing image quality metrics and human-opinion scores, concluding which metrics are best for quantifying perceptual quality.  In Fig.~\ref{fig:corrMethods}, we plot the mean-opinion scores of the methods included in the human-opinion study vs. the mean score according to the common full-reference measures RMSE, SSIM \cite{wang2004image}, IFC \cite{sheikh2005information}, and LPIPS \cite{zhang2018unreasonable}, as well as the no-reference methods by Ma et al.~\cite{ma2017learning}, NIQE \cite{mittal2013making}, BRISQUE \cite{mittal2012no} and the PI defined by~\eqref{eq:PI}. 
For each measure, we report Spearman's correlation coefficient with the raters' mean opinion scores, and also plot the corresponding least-squares linear fit.

As seen in Fig.~\ref{fig:corrMethods}, RMSE, SSIM and IFC, which are widely used for evaluating the quality of image reconstruction algorithms, are \emph{anti}-correlated with perceptual quality and thus inappropriate for evaluating it. Ma et al.~and BRISQUE show moderate correlation with human-opinion-scores, while LPIPS, NIQE and PI are highly correlated, with PI being the most correlated. 

The bottom pane of Fig.~\ref{fig:corrMethods} focuses on the high-perceptual quality regime, where it is important to distinguish between methods and correctly rank them. Metrics which excel in this regime will allow to assess progress in perceptual SR and to systematically compare methods. This is done by zooming in on the region of mean-opinion-score above $2.3$ (a new least-squares linear fit appears in magenta). These plots reveal that LPIPS, Ma et al.~and BRISQUE fail to faithfully quantify the perceptual quality in this regime. The only methods capable of correctly evaluating the perceptual quality of perceptually-aware SR algorithms are NIQE and PI (which is a combination of NIQE and Ma). Note that we also tested the full-reference measures VIF \cite{sheikh2006image}, FSIM \cite{zhang2011fsim} and MS-SSIM \cite{wang2003multiscale}, and the no-reference measures CORNIA \cite{ye2012unsupervised} and BLIINDS \cite{saad2012blind}, which all failed to correctly assess the perceptual quality\footnote{VIF, FSIM, MS-SSIM and CORNIA were \emph{anti}-correlated with the mean-opinion-scores. BLIINDS was moderately correlated, but failed in the high perceptual quality regime (similar to BRISQUE).}.

\begin{figure}[t]
	\begin{center}
		\includegraphics[width=0.99\linewidth]{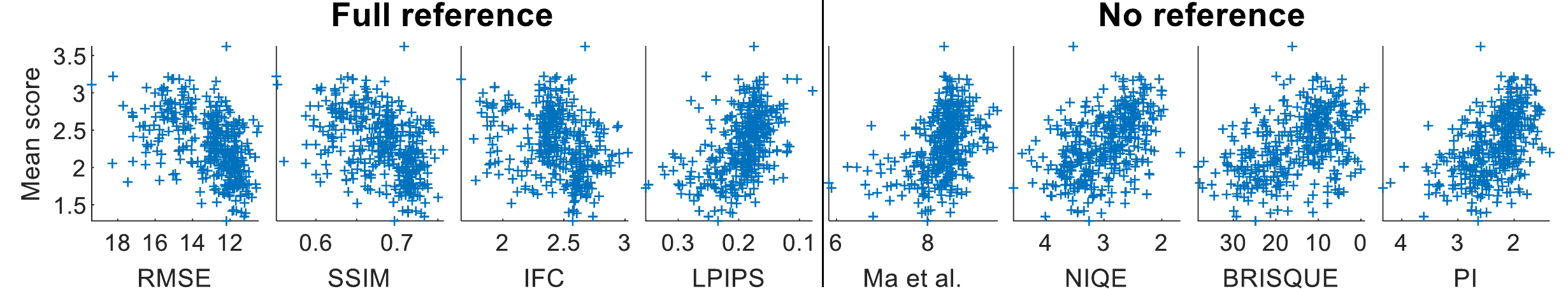}
		\caption{\textbf{Analysis of image quality measures on single images.} Scatter plots of $480$ outputs of challenge methods according to the mean-opinion-score ($y$-axis) and $8$ common image quality measures ($x$-axis). As above, RMSE, SSIM and IFC are \emph{anti}-correlated with human-opinion-scores, while NIQE and PI are most correlated (especially in the high perceptual quality regime).}\label{fig:corrImages}
	\end{center}
\end{figure}

We also analyze the correlation between human-opinion scores and common image quality measures on a single image. In Fig.~\ref{fig:corrImages} we plot the scores for outputs of each tested challenge method on all $40$ tested images ($480$ images altogether), where we average only over different human raters. To eliminate the variations between images (see Section \ref{sec:variability}), we first subtract the mean score of each image (over different raters) for both the human-opinion scores and the image quality measures. As can be seen, theses results are similar in trend to the results presented in Fig.~\ref{fig:corrMethods}.

%% file: currentTrends.tex
\section{Current Trends in Perceptual Super Resolution}\label{sec:trends}
All twenty-one groups who participated in the PIRM SR challenge, submitted algorithms based on deep nets. We next shortly review the current trends reflected in the submitted algorithms, in terms of three main aspects: the loss functions, the architectures, and methods to traverse the perception-distortion tradeoff. Note that the scope of this paper is not to review the field of SR, but rather to summarize the leading trends in the PIRM SR challenge. Additional details on the submitted methods can be found in the PIRM workshop proceedings.

\subsection{Loss functions}
Traditionally, neural networks for single image SR are trained with $\ell_1\slash \ell_2$ norm objectives \cite{zhao2017loss,yang2018deep}. These training objectives have been shown to enhance the values of common image evaluation metrics, e.g.~PSNR, SSIM. In the PIRM perceptual SR challenge, the evaluation methodology assesses the \emph{perceptual} quality of algorithms, which is not necessarily always enhanced by $\ell_1\slash \ell_2$ objectives \cite{blau2017perception}. As a consequence, a variety of other loss functions were suggested. The main observed trend is the use of adversarial training~\cite{goodfellow2014generative} in order to learn the statistics of natural images and reconstruct realistic images. Most participants used the standard GAN loss \cite{goodfellow2014generative}. Others \cite{wang2018esrgan} used a recent adaptation to the standard GAN loss named Relativistic GAN~\cite{jolicoeur2018relativistic}, which emphasizes the relation between the fake and real examples by modifying the loss function. Vu et al.~\cite{vu2018perception} suggested to further improve the relativistic GAN by wrapping it with the focal loss \cite{lin2017focal} which intensifies difficult samples and depresses easy samples.

Training the network solely with an adversarial loss is not enough since affinity to the input (distortion) is also of importance. The clear solution is to combine the GAN loss with the $\ell_1\slash \ell_2$ loss and by that target both perceptual quality and distortion. However, it was shown in \cite{ledig2017photo,sajjadi2017enhancenet} that $\ell_1\slash \ell_2$ losses prevent the generation of textures, which are crucial for perceptual quality. To overcome this, challenge participants used loss functions which are considered more perceptual (capture semantics). The ``perceptual loss''~\cite{johnson2016perceptual} appeared in most submitted solutions, where participants chose different nets and layers for extracting deep-features. An alternative for the perceptual loss used by \cite{Navarrete2018multi} is the contextual loss~\cite{mechrez2018learning,mechrez2018contextual}, that encourages the reconstructed images to have the same statistics as of the high resolution ground-truth images. 

A different approach \cite{gondal2018unreasonable} that achieved high perceptual quality is transferring texture by training with the Gram loss~\cite{gatys2015texture}, and without adversarial training. These participants show that standard texture transfer can be further improved by controlling the process using homogeneous semantic regions.

Submissions also applied other distortion functions, including the MS-SSIM loss function to emphasize a more structural distortion goal, Discrete Cosine Transform (DCT) based loss function and L1 norm between image gradients \cite{cheon2018generative} which were suggested in order overcome the smoothing effect of the MSE loss. 

\subsection{Architecture}
The second crucial component of submissions is the network architecture. Overall, most participating teams adopted state-of-the-art architectures from successful PSNR-maximization based SR methods and replaced the loss function. The main trend is to use the EDSR network architecture~\cite{lim2017enhanced} for the generator and the SRGAN architecture~\cite{ledig2017photo} for the discriminator. Wang et al.~\cite{wang2018esrgan} suggested to replace the residual block of EDSR with the Residual-in-Residual Dense Block (RRDB), which combines multi-level residual networks and dense connections. RRDR enables the use of deeper models, and as a result, improves the recovered textures. Others used Deep Back-Projection Networks (DBPN)~\cite{haris2018deep}, Enhanced Upscale Modules (EUSR)~\cite{kim2018deep}, and Multi-Grid-Back-Projection (MGBP)~\cite{Navarrete2018multi}.


\subsection{Traversing the perception-distortion tradeoff}
The tradeoff between perceptual quality and distortion raises the question of how to control the compromise between these two objectives. The importance of this question is two-fold: first, the optimal working point along the perception-distortion curve is domain specific and moreover it is image specific. Second, it is hard to predict the final working point, especially when the full objective is complex and when adversarial training is incorporated. Below we elaborate on four possible solutions (see pros and cons in Table~\ref{tab:procons}):

\begin{enumerate}
\item Retrain the network for each working point. This can be done by modifying the magnitude of the loss terms (e.g.~adversarial and distortion losses).
\item Interpolate between output images of two pretrained networks (in the pixel domain). For example, by using soft thresholding~\cite{deng2018enhancing}.
\item Interpolate between the parameters of two networks with the same architecture but different loss. This allows to generate a third network that is easy to control (see \cite{wang2018esrgan} for details). 
\item Control the tradeoff with an additional network input. For example, \cite{Navarrete2018multi} added noise to the input in order to traverse along the curve by changing the noise level at test time. 
\end{enumerate}


\begin{table}[h]
	\centering
	\ra{1.3}
	\begin{tabu}{@{}cllll@{}}
    \toprule
		\textbf{Method} & \phantom{ab} &  \textbf{Pros} & \phantom{ab} & \textbf{Cons}\\
    \midrule
		1 && Each working point is optimized && \makecell[l]{Not efficient, hard to control,\\large number of working points}\\
		2 && Simple && Inferior results\\
		3 && \makecell[l]{Easy to control, removes artifacts\\while maintaining textures} && \makecell[l]{The optimality of the outputs\\is not guaranteed}\\
		4  && Easy to control, efficient && \makecell[l]{The optimality of the outputs\\is not guaranteed}\\
	\bottomrule
	\end{tabu}
\caption{Pros and cons of the suggested methods for controlling the compromise between perceptual quality and distortion.}
\label{tab:procons}
\end{table}

%% file: conclusions.tex
\section{Conclusions}

The 2018 PIRM challenge is the first benchmark for perceptual-quality driven SR algorithms. The novel evaluation methodology used in this challenge enabled the assessment and ranking of perceptual SR methods along-side with those which target PSNR maximization. With this evaluation scheme, we compared the submitted algorithms with existing baselines, which revealed that the proposed methods push forward this field's state-of-the-art. A thorough study of the capability of common image quality measures to capture the perceptual quality of images was conducted. This study exposed that most common image quality measures are inadequate of quantifying perceptual quality. 

We conclude this report by pointing to several challenges in the field of perceptual SR, which should be the focus of future work. While we have witnessed major improvements over the past several years, in challenging scenarios such as $4$x SR, the outputs of current methods are generally unrealistic to human observers. This highlights that there is still much to be done to achieve high-quality perceptual SR images. Most common image quality measures fail to quantify the perceptual quality of SR methods, and there is still much room for improvement in this essential task. Perceptual-quality driven algorithms have yet to appear for the real-world scenario of blind SR. The perceptual quality objective, which has gained much attention for the SR task, should also gain attention for other image restoration tasks e.g.~deblurring. Finally, since a tradeoff between reconstruction accuracy and perceptual quality exists, schemes for controlling the compromise between the two can lead to adaptive SR schemes. This may promote new ways of quantifying the performance of SR algorithms, for instance, by measuring the area-under-the-curve in the perception-distortion plane.

\subsubsection*{Acknowledgments}
The 2018 PIRM Challenge on Perceptual SR was sponsored by Huawei and Mediatek.

%% file: appendices.tex
\section{Participating teams}\label{ap:teamMembers}
\begin{table}
	\tiny
	\centering
	\ra{1.3}
	\begin{tabular}{@{}lclcl@{}}\toprule
		Team name & \phantom{a} & Affiliation & \phantom{a} & Team members\\
		\toprule
		AIM  		&& KAIST && Thang Vu, Tung Luu\\\hline
		BOE  		&& BOE Technology Group Co., Ltd.	&& \makecell[l]{Pablo Navarrete Michelini, Dan Zhu,\\Hanwen Liu}\\\hline
		CEERI-lab 	&& \makecell[l]{\sus{1} IIIT-H\\\sus{2} CSIR-CEERI} && \makecell[l]{Rudrabha Mukhopadhyay\sus{1}, Manoj Sharma\sus{2},\\ Utkarsh Verma\sus{2}, Shubham Jain\sus{2},\\Sagnik Bhowmick\sus{2}, Avinash Upadhyay\sus{2},\\Sriharsha Koundinya\sus{2}, Ankit Shukla\sus{2}}\\\hline
		CLFStudio 	&& \makecell[l]{\sus{1} East China Normal University\\\sus{2} Jiangxi Normal University}	&& \makecell[l]{Juncheng Li\sus{1}, Kangfu Mei\sus{2}, Faming Fang\sus{1},\\Yiting Yuan\sus{1}}\\\hline
		DSP-whu  	&& Wuhan University	&& Ye Yang, Sheng Tian, Yuhan Hu\\\hline
		gayNet  	&& -	&& YH Liu, ZP Zhang\\\hline
		Haiyun-xmu 	&& Xiamen university	&& \makecell[l]{Rong Chen, Xiaotong Luo, Yanyun Qu,\\Cuihua Li}\\\hline
		IPCV		&& \makecell[l]{Indian Institute of Technology, Madras, India}	&& \makecell[l]{Subeesh Vasu, Nimisha Thekke Madam,\\A.N. Rajagopalan}\\\hline
		Yonsei-MCML && Yonsei University	&& \makecell[l]{Jun-Hyuk Kim, Jun-Ho Choi, Manri Cheon,\\Jong-Seok Lee}\\\hline
		PDSR 		&& Duke University	&& Alina Jade Barnett, Lei Chen, Cynthia Rudin\\\hline
		REC-SR 		&& \makecell[l]{Indian Institute of Technology, Madras, India}	&& \makecell[l]{Kuldeep Purohit, Srimanta Mandal,\\A.N. Rajagopalan}\\\hline
		SI Analytics && Satrec Initiative	&& Junghoon Seo, SeungHyun Jeon\\\hline
		SMILE 		&& Northeastern University && \makecell[l]{Yulun Zhang, Kunpeng Li, Kai Li,\\Lichen Wang, Bineng Zhong, Yun Fu}\\\hline
		SuperSR 	&& 	\makecell[l]{\sus{1} The Chinese University of Hong Kong\\\sus{2} Shenzhen Institutes of Advanced Technology\\\sus{3} The Chinese University of Hong Kong, Shenzhen\\\sus{4} Nanyang Technological University, Singapore\\}	&& 	\makecell[l]{Xintao Wang\sus{1}, Shixiang Wu\sus{2}, Jinjin Gu\sus{3},\\Ke Yu\sus{1}, Yihao Liu\sus{2}, Chao Dong\sus{2}, Yu Qiao\sus{2},\\Chen Change Loy\sus{4}}	\\\hline
		TSRN 		&& \makecell[l]{\sus{1} Max Planck Institute for Intelligent Systems\\\sus{2} Amazon Research}	&& \makecell[l]{Muhammad Waleed Gondal\sus{1},\\Bernhard Schoelkopf\sus{1}, Michael Hirsch\sus{2}} \\\hline
		TTI 		&& \makecell[l]{\sus{1} Toyota Technological Institute\\\sus{2} Toyota Technological Institute at Chicago}	&& \makecell[l]{Muhammad Haris\sus{1}, Tomoki Yoshida\sus{1},\\Kazutoshi Akita\sus{1}, Norimichi Ukita\sus{1},\\Greg Shakhnarovich\sus{2}}		\\\hline
		VIPSL 		&& Xidian University	&& \makecell[l]{Yuanfei Huang, Ruihan Dou, Furui Bai,\\Rui Wang, Wen Lu, Xinbo Gao}		\\\hline
		XYN 		&& 	Wuhan University	&& Sheng Tian, Ye Yang, Yuhan HU, Yuan Fu	\\\hline
		ZY.FZU 		&& \makecell[l]{\sus{1} Fuzhou University\\\sus{2} Imperial Vision Technology} && \makecell[l]{Yan Zhao\sus{1}, Kehui Nie\sus{1}, Gen Li\sus{2},\\Qinquan Gao\sus{1}}\\
		\bottomrule
	\end{tabular}
	\caption{Participating teams (alphabetical order).}\label{tab:team}
\end{table}

\clearpage
\section{Test phase results}\label{app:fullResults}
\begin{table}
	\scriptsize
	\centering
	\ra{1.3}
	\begin{tabular}{@{}llcccllcccllcc@{}}\toprule
		\multicolumn{4}{c}{\textbf{Region 1}} & \phantom{ab}& \multicolumn{4}{c}{\textbf{Region 2}} & \phantom{ab} & \multicolumn{4}{c}{\textbf{Region 3}}\\
		\cmidrule{1	- 4} \cmidrule{6 - 9} \cmidrule{11 - 14} \# & Team & \makecell{PI} & RMSE && \# & Team & \makecell{PI} & RMSE && \# &Team & \makecell{PI} & RMSE\\
		\cmidrule{1	- 4} \cmidrule{6 - 9} \cmidrule{11 - 14}
		$1$  & IPCV		 	& 2.709 & 11.48 && $1$ 	& TTI			& 2.199	& 12.40	&& $1$	& SuperSR 		& 1.978 & 15.30\\
		$2$  & Yonsei-MCML	& 2.750	& 11.44	&& $2*$ & IPCV-team		& 2.275 & 12.47	&& $2$  & BOE			& 2.019	& 14.24\\
		$3*$ & SuperSR		& 2.933	& 11.50	&& $2*$ & Yonsei-MCML	& 2.279 & 12.41 && $3$  & IPCV-team		& 2.013	& 15.26\\
		$3*$ & TTI			& 2.938	& 11.46	&& $4$ 	& SuperSR		& 2.424	& 12.50	&& $4$  & AIM			& 2.013	& 15.60\\
		$5$  & AIM			& 3.321	& 11.37	&& $5$ 	& BOE			& 2.484 & 12.50	&& $5$  & TTI			& 2.040	& 13.17\\
		$6$  & DSP-whu		& 3.728	& 11.45	&& $6$ 	& AIM			& 2.600 & 12.42	&& $6$  & Haiyun-xmu	& 2.077	& 15.95\\
		$7*$ & BOE			& 3.817	& 11.50	&& $7$ 	& REC-SR		& 2.635 & 12.37	&& $7$  & gayNet		& 2.104	& 15.88\\
		$7*$ & REC-SR		& 3.831	& 11.46	&& $8$ 	& DSP-whu		& 2.660 & 12.24 && $8$  & DSP-whu		& 2.114	& 15.93\\
		$9$  & Haiyun-xmu	& 4.440	& 11.19	&& $9$ 	& XYN			& 2.946 & 12.23	&& $9$  & Yonsei-MCML	& 2.136	& 13.44\\
		$10$ & PDSR			& 4.818	& 10.70	&& 		&				& 	 	& 		&& $10$ & REC-SR		& 2.126	& 14.85\\
		$11$ & SMILE		& 5.034	& 10.59	&& 		&				&  		& 		&& $11$ & XYN			& 2.164	& 15.73\\
		$12$ & CLFStudio	& 5.244	& 11.47	&& 		&				&  		& 		&& $12$ & TSRN			& 2.227	& 15.66\\
		$13$ & CEERI-lab	& 5.890	& 11.46	&& 		&				& 		& 		&& $13$ & SI Analytics	& 2.295	& 14.91\\
		& 					& 		& 		&& 		&				& 		&		&& $14$ & ZY.FZU		& 2.387	& 14.75\\
		& 					& 		& 		&& 		&				&		& 		&& $15$ & SMILE			& 2.405	& 13.85\\
		& 					& 		& 		&& 		&				& 		&		&& $16$ & Try-Me		& 2.441	& 13.35\\
		&					& 		& 		&& 		&				& 		&		&& $17$ & VIPSL			& 2.452	& 14.60\\
		& 					&		& 		&& 		&				& 		&		&& $18$ & ILC			& 2.594	& 12.53\\
		\bottomrule
	\end{tabular}
	\caption{\textbf{Challenge results.} The top submission of each group in each region. For submissions with a marginal perceptual index difference (up to 0.01), the one with the lower RMSE is ranked higher. Submission with marginal differences in both the perceptual index and RMSE are ranked together (marked by *).}\label{tab:fullResults}
\end{table}

\clearpage
\section{More results}\label{app:moreResults}
\begin{figure}[h!]
	\begin{center}
		\includegraphics[width=0.78\linewidth]{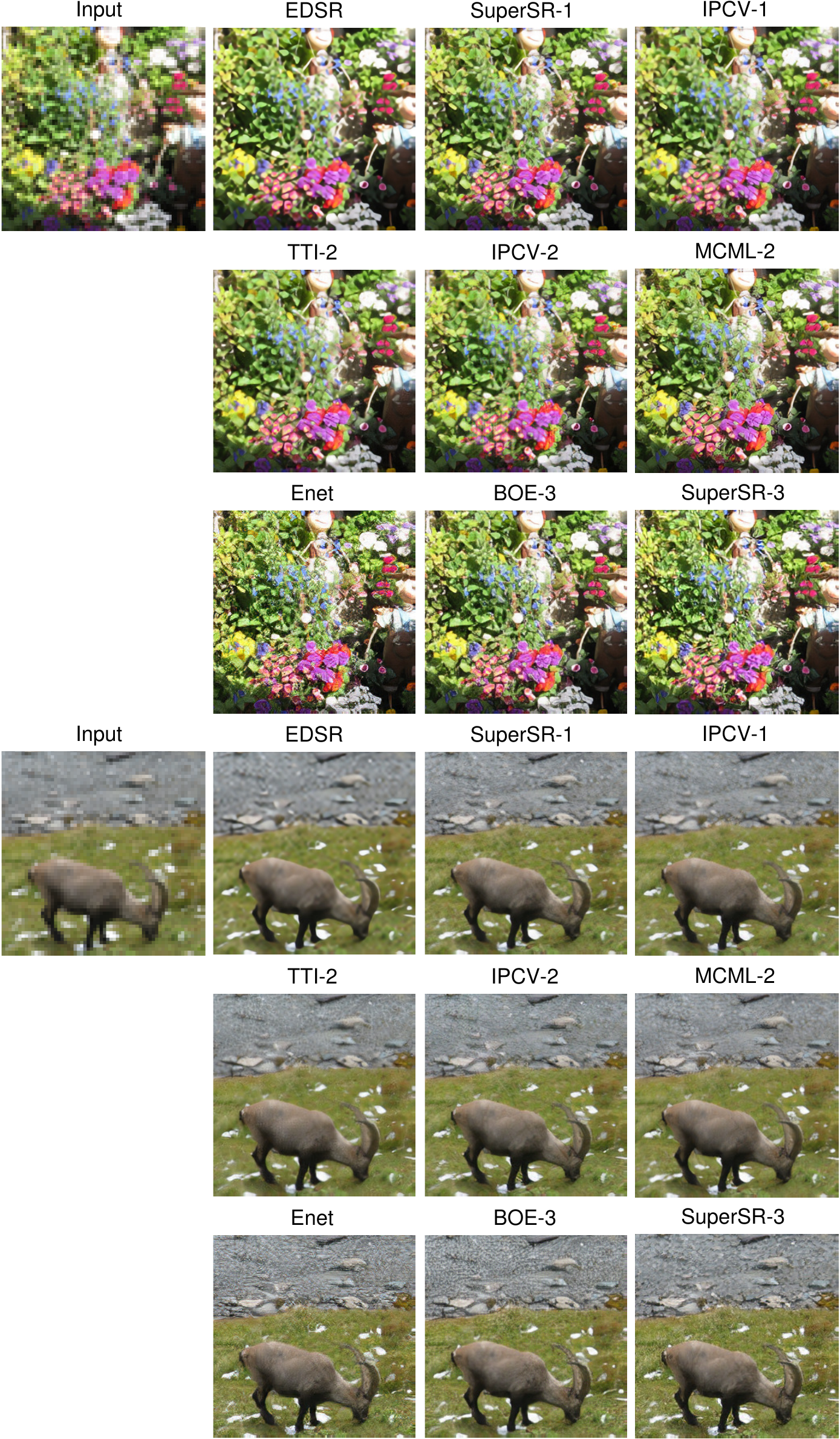}
		\caption{\textbf{Visual results.} Additional SR results of several top methods in each region, along with baselines \cite{lim2017enhanced,sajjadi2017enhancenet}.}\label{fig:topImages2}
	\end{center}
\end{figure}
\begin{figure}
	\begin{center}
		\includegraphics[width=0.78\linewidth]{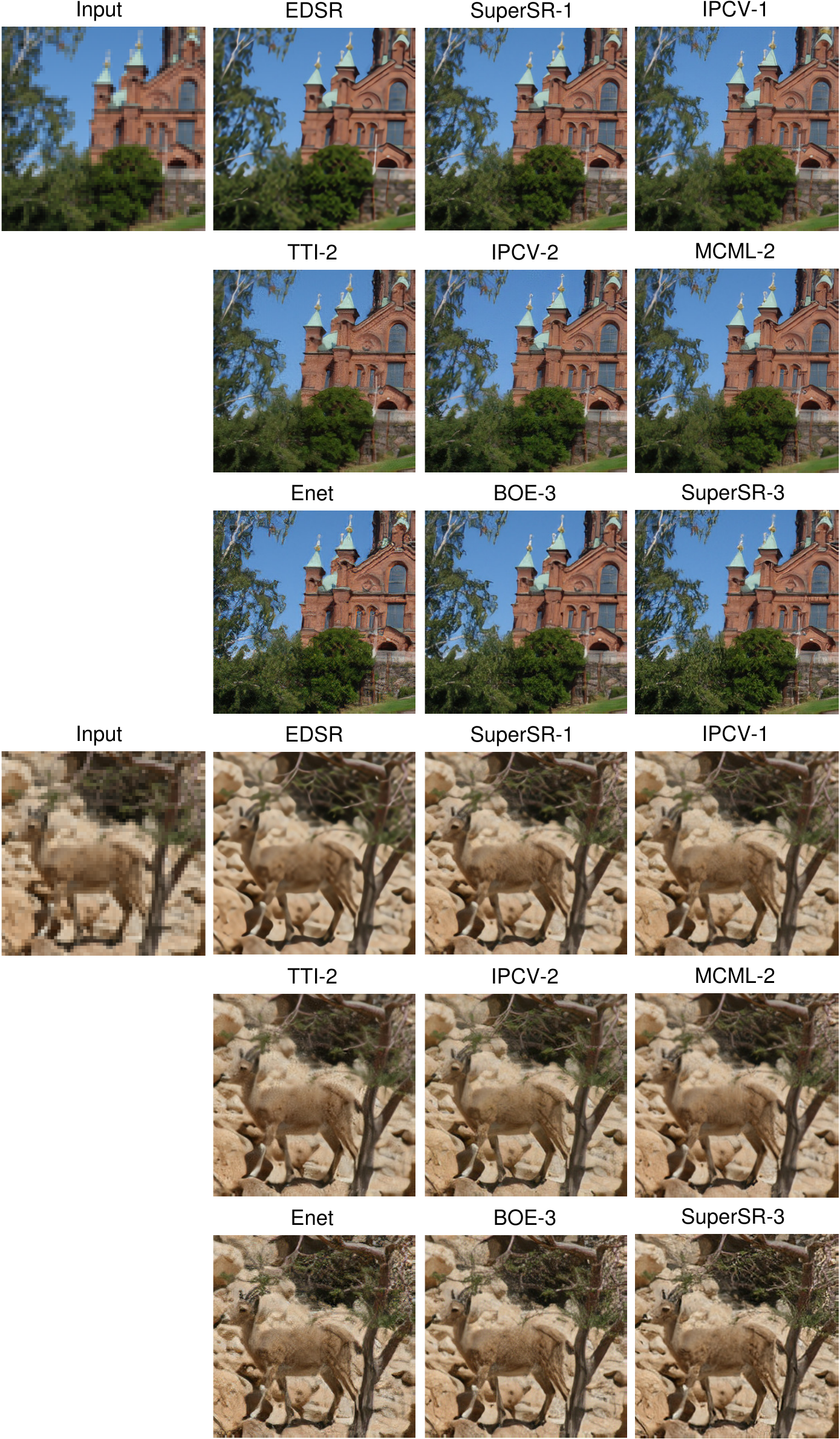}
		\caption{\textbf{Visual results.} Additional SR results of several top methods in each region, along with baselines \cite{lim2017enhanced,sajjadi2017enhancenet}.}\label{fig:topImages3}
	\end{center}
\end{figure}

%% file: main.bbl
\begin{thebibliography}{10}
\providecommand{\url}[1]{\texttt{#1}}
\providecommand{\urlprefix}{URL }
\providecommand{\doi}[1]{https://doi.org/#1}

\bibitem{blau2017perception}
Blau, Y., Michaeli, T.: The perception-distortion tradeoff. In: Proc. CVPR
  (2018)

\bibitem{cheon2018generative}
Cheon, M., Kim, J.H., Choi, J.H., Lee, J.S.: Generative adversarial
  network-based image super-resolution using perceptual content losses. In:
  Proc. ECCV Workshops (2018)

\bibitem{choi2018deep}
Choi, J.H., Kim, J.H., Cheon, M., Lee, J.S.: Deep learning-based image
  super-resolution considering quantitative and perceptual quality. arXiv
  preprint arXiv:1809.04789  (2018)

\bibitem{dahl2017pixel}
Dahl, R., Norouzi, M., Shlens, J.: Pixel recursive super resolution. In: Proc.
  ICCV (2017)

\bibitem{deng2018enhancing}
Deng, X.: Enhancing image quality via style transfer for single image
  super-resolution. IEEE Signal Processing Letters  \textbf{25}(4),  571--575
  (2018)

\bibitem{dong2014learning}
Dong, C., Loy, C.C., He, K., Tang, X.: Learning a deep convolutional network
  for image super-resolution. In: Proc. ECCV (2014)

\bibitem{gatys2015texture}
Gatys, L., Ecker, A.S., Bethge, M.: Texture synthesis using convolutional
  neural networks. In: Proc. NIPS (2015)

\bibitem{gondal2018unreasonable}
Gondal, M.W., Sch{\"o}lkopf, B., Hirsch, M.: The unreasonable effectiveness of
  texture transfer for single image super-resolution. In: Proc. ECCV Workshops
  (2018)

\bibitem{goodfellow2014generative}
Goodfellow, I., Pouget-Abadie, J., Mirza, M., Xu, B., Warde-Farley, D., Ozair,
  S., Courville, A., Bengio, Y.: Generative adversarial nets. In: Proc. NIPS
  (2014)

\bibitem{han2018image}
Han, W., Chang, S., Liu, D., Yu, M., Witbrock, M., Huang, T.S.: Image
  super-resolution via dual-state recurrent networks. In: Proc. CVPR (2018)

\bibitem{haris2018deep}
Haris, M., Shakhnarovich, G., Ukita, N.: Deep backprojection networks for
  super-resolution. In: Proc. CVPR (2018)

\bibitem{huang2015single}
Huang, J.B., Singh, A., Ahuja, N.: Single image super-resolution from
  transformed self-exemplars. In: Proc. CVPR (2015)

\bibitem{johnson2016perceptual}
Johnson, J., Alahi, A., Fei-Fei, L.: Perceptual losses for real-time style
  transfer and super-resolution. In: Proc. ECCV (2016)

\bibitem{jolicoeur2018relativistic}
Jolicoeur-Martineau, A.: The relativistic discriminator: a key element missing
  from standard {GAN}. arXiv preprint arXiv:1807.00734  (2018)

\bibitem{kim2016accurate}
Kim, J., Kwon~Lee, J., Mu~Lee, K.: Accurate image super-resolution using very
  deep convolutional networks. In: Proc. CVPR (2016)

\bibitem{kim2018deep}
Kim, J.H., Lee, J.S.: Deep residual network with enhanced upscaling module for
  super-resolution. In: Proc. CVPR Workshops (2018)

\bibitem{lai2017deep}
Lai, W.S., Huang, J.B., Ahuja, N., Yang, M.H.: Deep laplacian pyramid networks
  for fast and accurate superresolution. In: Proc. CVPR (2017)

\bibitem{ledig2017photo}
Ledig, C., Theis, L., Husz{\'a}r, F., Caballero, J., Cunningham, A., Acosta,
  A., Aitken, A.P., Tejani, A., Totz, J., Wang, Z., et~al.: Photo-realistic
  single image super-resolution using a generative adversarial network. In:
  Proc. CVPR (2017)

\bibitem{lim2017enhanced}
Lim, B., Son, S., Kim, H., Nah, S., Lee, K.M.: Enhanced deep residual networks
  for single image super-resolution. In: Proc. CVPR workshops (2017)

\bibitem{lin2017focal}
Lin, T.Y., Goyal, P., Girshick, R., He, K., Doll{\'a}r, P.: Focal loss for
  dense object detection. In: Proc. ICCV (2017)

\bibitem{luo2018bi}
Luo, X., Chen, R., Xie, Y., Qu, Y., Cui-hua, L.: {Bi-GANs-ST} for perceptual
  image super-resolution. In: Proc. ECCV Workshops (2018)

\bibitem{ma2017learning}
Ma, C., Yang, C.Y., Yang, X., Yang, M.H.: Learning a no-reference quality
  metric for single-image super-resolution. Computer Vision and Image
  Understanding  \textbf{158},  1--16 (2017)

\bibitem{martin2001database}
Martin, D., Fowlkes, C., Tal, D., Malik, J.: A database of human segmented
  natural images and its application to evaluating segmentation algorithms and
  measuring ecological statistics. In: Proc. ICCV (2001)

\bibitem{mechrez2018learning}
Mechrez, R., Talmi, I., Shama, F., Zelnik-Manor, L.: Learning to maintain
  natural image statistics. arXiv preprint arXiv:1803.04626  (2018)

\bibitem{mechrez2018contextual}
Mechrez, R., Talmi, I., Zelnik-Manor, L.: The contextual loss for image
  transformation with non-aligned data. In: Proc. ECCV (2018)

\bibitem{mittal2012no}
Mittal, A., Moorthy, A.K., Bovik, A.C.: No-reference image quality assessment
  in the spatial domain. IEEE Transactions on Image Processing (TIP)
  \textbf{21}(12),  4695--4708 (2012)

\bibitem{mittal2013making}
Mittal, A., Soundararajan, R., Bovik, A.C.: Making a ``completely blind'' image
  quality analyzer. IEEE Signal Process. Lett.  \textbf{20}(3),  209--212
  (2013)

\bibitem{Navarrete2018multi}
Navarrete~Michelini, P., Zhu, D., Hanwen, L.: Multi-scale recursive and
  perception-distortion controllable image super-resolution. In: Proc. ECCV
  Workshops (2018)

\bibitem{kuldeep2018scale}
Purohit, K., Mandal, S., Rajagopalan, A.N.: Scale-recurrent multi-residual
  dense network for image super resolution. In: Proc. ECCV Workshops (2018)

\bibitem{saad2012blind}
Saad, M.A., Bovik, A.C., Charrier, C.: Blind image quality assessment: A
  natural scene statistics approach in the {DCT} domain. IEEE transactions on
  Image Processing (TIP)  \textbf{21}(8),  3339--3352 (2012)

\bibitem{sajjadi2017enhancenet}
Sajjadi, M.S., Sch{\"o}lkopf, B., Hirsch, M.: Enhancenet: Single image
  super-resolution through automated texture synthesis. In: Proc. ICCV (2017)

\bibitem{sheikh2006image}
Sheikh, H.R., Bovik, A.C.: Image information and visual quality. IEEE
  Transactions on image processing (TIP)  \textbf{15}(2),  430--444 (2006)

\bibitem{sheikh2005information}
Sheikh, H.R., Bovik, A.C., De~Veciana, G.: An information fidelity criterion
  for image quality assessment using natural scene statistics. IEEE
  Transactions on image processing  \textbf{14}(12),  2117--2128 (2005)

\bibitem{shocher2017zero}
Shocher, A., Cohen, N., Irani, M.: ``zero-shot'' super-resolution using deep
  internal learning. In: Proc. CVPR (2018)

\bibitem{sun2017super}
Sun, L., Hays, J.: Super-resolution using constrained deep texture synthesis.
  arXiv preprint arXiv:1701.07604  (2017)

\bibitem{timofte2017ntire}
Timofte, R., Agustsson, E., Van~Gool, L., Yang, M.H., Zhang, L., et~al.:
  {NTIRE} 2017 challenge on single image super-resolution: Methods and results.
  In: Proc. CVPR workshops (2017)

\bibitem{timofte2014a+}
Timofte, R., De~Smet, V., Van~Gool, L.: A+: Adjusted anchored neighborhood
  regression for fast super-resolution. In: Proc. ACCV (2014)

\bibitem{timofte2018ntire}
Timofte, R., Gu, S., Wu, J., Van~Gool, L., Zhang, L., Yang, M.H., et~al.:
  {NTIRE} 2018 challenge on single image super-resolution: Methods and results.
  In: Proc. CVPR workshops (2018)

\bibitem{tong2017image}
Tong, T., Li, G., Liu, X., Gao, Q.: Image super-resolution using dense skip
  connections. In: Proc. ICCV (2017)

\bibitem{vasu2018analyzing}
Vasu, S., Nimisha, T.M., Rajagopalan, A.N.: Analyzing perception-distortion
  tradeoff using enhanced perceptual super-resolution network. In: Proc. ECCV
  Workshops (2018)

\bibitem{vu2018perception}
Vu, T., Luu, T., Yoo, C.D.: Perception-enhanced image super-resolution via
  relativistic generative adversarial networks. In: Proc. ECCV Workshops (2018)

\bibitem{wang2018recovering}
Wang, X., Yu, K., Dong, C., Loy, C.C.: Recovering realistic texture in image
  super-resolution by deep spatial feature transform. In: Proc. CVPR (2018)

\bibitem{wang2018esrgan}
Wang, X., Yu, K., Wu, S., Gu, J., Liu, Y., Dong, C., Qiao, Y., Loy, C.C.:
  {ESRGAN}: Enhanced super-resolution generative adversarial networks. In:
  Proc. ECCV Workshops (2018)

\bibitem{wang2018fully}
Wang, Y., Perazzi, F., McWilliams, B., Sorkine-Hornung, A., Sorkine-Hornung,
  O., Schroers, C.: A fully progressive approach to single-image
  super-resolution. In: Proc. CVPR (2018)

\bibitem{wang2004image}
Wang, Z., Bovik, A.C., Sheikh, H.R., Simoncelli, E.P.: Image quality
  assessment: from error visibility to structural similarity. IEEE transactions
  on image processing (TIP)  \textbf{13}(4),  600--612 (2004)

\bibitem{wang2003multiscale}
Wang, Z., Simoncelli, E.P., Bovik, A.C.: Multiscale structural similarity for
  image quality assessment. In: Conference on Signals, Systems \& Computers.
  vol.~2, pp. 1398--1402 (2003)

\bibitem{yang2018deep}
Yang, W., Zhang, X., Tian, Y., Wang, W., Xue, J.H.: Deep learning for single
  image super-resolution: A brief review. arXiv preprint arXiv:1808.03344
  (2018)

\bibitem{ye2012unsupervised}
Ye, P., Kumar, J., Kang, L., Doermann, D.: Unsupervised feature learning
  framework for no-reference image quality assessment. In: Proc. CVPR (2012)

\bibitem{zhang2011fsim}
Zhang, L., Zhang, L., Mou, X., Zhang, D., et~al.: {FSIM}: a feature similarity
  index for image quality assessment. IEEE transactions on Image Processing
  (TIP)  \textbf{20}(8),  2378--2386 (2011)

\bibitem{zhang2018unreasonable}
Zhang, R., Isola, P., Efros, A.A., Shechtman, E., Wang, O.: The unreasonable
  effectiveness of deep features as a perceptual metric. In: Proc. CVPR (2018)

\bibitem{zhang2018image}
Zhang, Y., Li, K., Li, K., Wang, L., Zhong, B., Fu, Y.: Image super-resolution
  using very deep residual channel attention networks. In: Proc. ECCV (2018)

\bibitem{zhang2018residual}
Zhang, Y., Tian, Y., Kong, Y., Zhong, B., Fu, Y.: Residual dense network for
  image super-resolution. In: Proc. CVPR (2018)

\bibitem{zhao2017loss}
Zhao, H., Gallo, O., Frosio, I., Kautz, J.: Loss functions for image
  restoration with neural networks. IEEE Transactions on Computational Imaging
  \textbf{3}(1),  47--57 (2017)

\end{thebibliography}
